# Evolving Thematic Map Design in Academic Cartography: A Thirty-Year Study Based on Multilingual Journals

Zhiwei WEI[a,b], Chenxi SONG[a,b], Tazhu WANG[a,b], Fan WU[c], Hua LIAO[a,b], Su DING[d], Nai YANG[c]

[a]*Hunan Normal University, School of Geographic Sciences, Hunan Changsha, China;* [b]*Hunan Key Laboratory of Geospatial Big Data Mining and Application, Hunan Changsha, China*; [c]*School of Geography and Information Engineering, China University of Geosciences, Wuhan, China*; [d]*School of Environmental and Resources Science, Zhejiang A&F University, Hangzhou, China.*

*Corresponding Author: Nai YANG. E-mail:yangnai@cug.edu.cn

# Evolving Thematic Map Design in Academic Cartography: A Thirty-Year Study Based on Multilingual Journals

**Abstract**: Thematic maps play a central role in academic communication, yet their large-scale design evolution has rarely been examined empirically. This study presents a longitudinal and multilingual analysis of thematic map design practices in academic cartography from 1990 to 2020. We compile a corpus of 45,732 research articles from sixteen authoritative Chinese- and English-language journals and extract 23,928 maps using computer vision and large-model-based document parsing to build a structured dataset. Map design characteristics are quantified across three dimensions: map elements, color design, and layout structure. Results show that Chinese- and English-language academic maps share highly similar structural conventions, typically employing restrained color palettes with neutral dominant hues, low saturation, high brightness, and limited hue diversity, as well as centered layouts with high main-map occupation ratios. Differences exist in that English-language maps show slightly greater hue richness and compactness, whereas Chinese-language maps historically rely more on neutral hues and integrated layouts. Temporal analysis reveals parallel evolutionary trends in both groups, including increasing element richness, legend usage, and hue diversity, alongside stable layout structures. Overall, the findings suggest that academic map design evolution is characterized more by institutional convergence than cultural divergence.

## 1. Introduction

Maps have long served as a key medium for communicating geographic knowledge in academic research (White et al., 2017; Coetzee et al., 2021). As scientific publishing continues to expand at an unprecedented pace, cartographic visualizations in academic journals have become increasingly critical for presenting complex spatial analyses and engaging readers (Wu et al., 2024). Reflecting this shift, many journals now place greater emphasis on the visual quality and clarity of figures, and some have introduced more explicit guidelines or professional support for figure and map preparation (Wang et al., 2024). Consequently, design choices related to map elements, color usage, and layout structure have a direct impact on how scientific arguments are perceived and understood.

A substantial body of research in cartography has investigated map design principles, offering theoretical frameworks and practical guidance. In terms of map elements, prior studies have examined the selection and visual hierarchy of core components such as the main map and legend, emphasizing their role in conveying structure and importance (Kessler & Slocum, 2011). More recently, Jia et al. (2023) have summarized the structural elements of the ubiquitous map. As for color design, a comprehensive set of guidelines grounded in the theory of visual variables was first proposed by Bertin (1974). Then the subsequent works have explored how hue, brightness, and contrast influence perception, readability, and semantic interpretation, leading to widely adopted guidelines for thematic mapping and visual differentiation (e.g., Brewer, 1994; MacEachren, 20024; Silva et al., 2011; Yang et al., 2025). With respect to layout structure, studies have addressed the spatial organization of the main map and auxiliary elements, such as legends and titles, highlighting principles of balance, alignment, and visual flow in page-based map composition (e.g., Dent et al., 2008; Slocum et al., 2009). Together, these works form a well-established foundation for understanding cartographic design. However, much of this knowledge has been derived from expert reasoning, instructional examples, or analyses of relatively small collections of maps.

Recently, addressing the above experience-based and small-sample nature of existing map-related studies has become feasible. The widespread adoption of open-access publishing, together with advances in document analysis and machine learning, has made it increasingly feasible to systematically collect academic maps and automatically extract and analyze their visual components at scale (Zhang et al., 2025). For example, some researchers have focused on scanned map images and thematic content recognition, such as map-type classification and text extraction (e.g., Mandal et al., 2014; Zhou et al., 2018). Subsequent studies expanded toward generative and segmentation-oriented tasks, including image-to-image translation between map styles (Isola et al., 2017) and feature extraction from topographic or historical maps (Saeedimoghaddam & Stepinski, 2020; Kramm et al., 2025). More recent datasets have begun to consider maps as structured visual documents by annotating cartographic elements and layout components, exemplified by CartoMark (Zhou et al., 2024), ubiMap and ubiMap-l (Yang et al., 2025). Despite these advances, most existing datasets remain task-driven and are designed to support visual recognition, segmentation, or text detection, with limited attention to cartographic design principles and their empirical evaluation. As a result, large-scale, longitudinal analyses that explicitly target map design characteristics, such as color usage and layout structure, remain largely unexplored.

Motivated by these considerations, this study tries to investigate map design through large-scale empirical evidence drawn from academic publications. We compile a corpus of 45,732 research articles from 16 authoritative cartography and geographic information science journals, comprising eight Chinese-language journals and eight English-language journals, published between 1990 and 2020. Based on this corpus, we construct a large-scale academic map dataset (23928 maps) by automatically identifying and extracting maps from full-text articles, and further isolating the main map area and key map elements using large-model-based document understanding and computer vision techniques (*Section* 3). Building upon these extracted components, we examine academic map design along three analytical dimensions: map elements, color design, and layout structure (*Section* 4). The analysis focuses on two interconnected objectives: identifying cross-language differences between Chinese- and English-language journals and tracing temporal evolution over three decades (*Section* 5). The overall analytical framework of the study is illustrated in Figure 1.

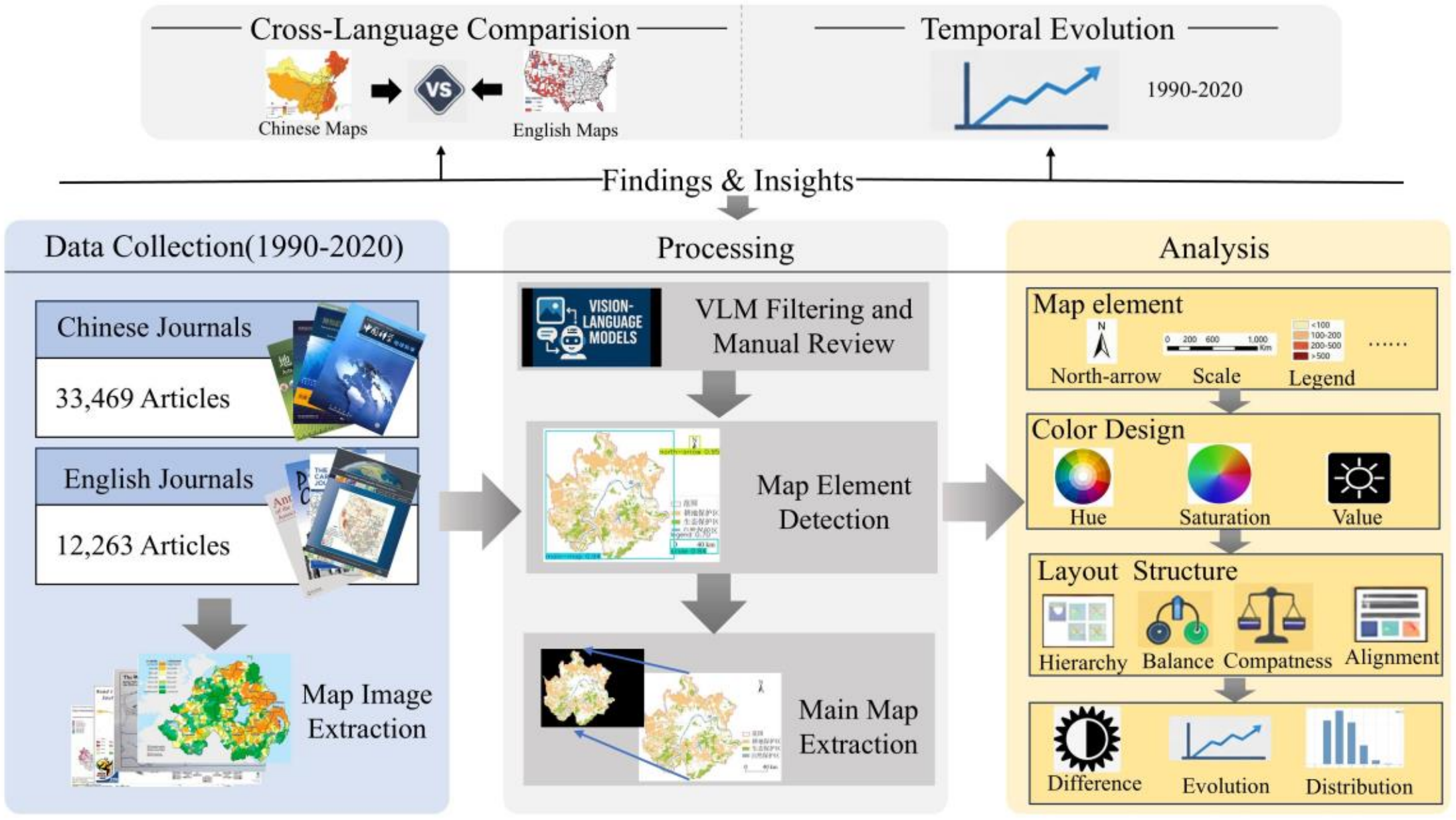

Figure 1. The overall analytical framework of this study.

# 2. Related works

## 2.1Map color and layout design

Map design has been extensively studied in cartography, resulting in a broad body of research on visual and structural design principles. Given the scope of this work, we selectively review representative studies and focus on (1) **color design** and (2) **layout structure**, which are most relevant to our analysis. Map elements are not discussed separately, as they are typically considered integral components of layout structure in both theoretical and empirical studies.

**(1) Color design**

Color design is one of the most thoroughly studied aspects of map design, as color plays a critical role in encoding thematic information and guiding visual perception. Early foundational work by Bertin (1974) introduced the theory of visual variables, identifying hue, value (brightness), and saturation as fundamental dimensions for visual encoding. This framework has profoundly influenced subsequent cartographic research and remains a cornerstone of color design theory. Building on Bertin's work, MacEachren (2004) further connected visual variables to cognitive and perceptual processes, emphasizing the alignment between color choices, data characteristics, and map-reading tasks. Building on these theoretical foundations, researchers have sought to translate color theory into practical guidelines for map design. Brewer (1994) systematically investigated color use in maps and proposed practical guidelines that relate color schemes to data measurement levels and mapping purposes. These principles were later operationalized through widely adopted tools and templates, such as ColorBrewer (Brewer et al., 2003; Harrower & Brewer, 2003), which summarize effective qualitative, sequential, and diverging color schemes for cartographic applications. Subsequent studies have examined perceptual aspects of color design, including brightness contrast, class discriminability, and visual consistency, demonstrating their influence on map readability and user interpretation (Brewer et al., 1997; Brychtová & Çöltekin, 2015; Ware, 2019). Beyond general guidelines, research has also emphasized semantic conventions and contextual appropriateness in map color design. Conventional color associations (e.g., blue for water, green for vegetation) and adherence to cartographic or institutional standards have been shown to reduce cognitive load and support intuitive interpretation. More recent work has begun to formalize color knowledge through rule-based systems and interactive tools, integrating insights from perception, cartography, and design practice (Christophe, 2011). Yang et al. (2025) even used this knowledge for a large language model to support interactive color design in practice. Together, these studies establish color design as a mature and theoretically grounded area of cartographic research, while also highlighting the diversity of design choices observed in practice.

**(2) Layout structure**

Layout design addresses the spatial organization of the main map and auxiliary elements, such as legends, titles, scale bars, and inset maps, within a bounded page or screen (Jia et al., 2024). Classical cartographic literature has long emphasized layout principles derived from visual perception and graphic design, including visual hierarchy, balance, alignment, and Gestalt grouping (Dent et al., 2008; Slocum et al., 2009; MacEachren, 2004). These principles aim to ensure that the main map remains the dominant visual focus while supporting elements are arranged to facilitate efficient information access without competing for attention. In addition to the above qualitative guidelines, researchers have increasingly sought to formalize and quantify layout quality. Early efforts modeled visual balance using mechanical analogies or geometric measures (Ma et al., 2013), while later studies proposed metrics for hierarchy, compactness, and alignment to evaluate layout effectiveness (O' Donovan et al., 2014). On the generative side, several approaches have attempted to translate expert layout knowledge into computational rules or optimization procedures, including case-based recommendation systems (Wei et al., 2017), explicit legend grouping rules (Qin & Li, 2017), and integer programming-based layout optimization (Dayama et al., 2020). More recently, data-driven approaches have gained prominence in layout research. Inspired by advances in document and interface design, learning-based models have been applied to capture layout patterns from large-scale datasets, such as LayoutGAN and LayoutVAE (Li et al., 2019; Jyothi et al., 2019). Within cartography, recent studies have begun to move beyond rule-based or task-specific formulations and explore data-driven representations of map layouts. MapLayNet (Yang et al., 2025) represents an early effort to learn map layout representations in a data-driven manner by modeling map elements and their spatial relationships as graphs and employing weakly supervised graph neural networks to capture global-local layout patterns from large collections of ubiquitous maps.

In summary, prior research has established comprehensive design principles for both color usage and layout organization, supported by decades of theoretical development and empirical investigation. However, most existing studies focus on prescribing design guidelines, developing tools, or analyzing limited collections of exemplary maps. Large-scale, longitudinal analyses that examine how these design principles are applied in academic publishing remain scarce.

## 2.2 Data-driven analysis of maps

With the rapid expansion of digital mapping resources and the development of intelligent analysis techniques, conducting large-scale data-driven studies of maps has become increasingly feasible in recent years (Robinson et al., 2017). Geospatial artificial intelligence (GeoAI) research has demonstrated that deep learning and related methods can effectively support large-scale geographic image analysis and mapping tasks by leveraging abundant digital map and spatial data sources (Li & Hsu, 2022).

Early data-driven studies on maps primarily treated map images as visual inputs for recognition-oriented tasks, focusing on scanned cartographic products and thematic classification. Representative datasets compiled annotated map images into coarse thematic categories, such as political, physical, or topographic maps, to support research on text extraction, symbol recognition, and map-type classification (Mandal et al., 2014). With the adoption of deep convolutional neural networks, subsequent datasets expanded in scale and were used to train models that classify maps into different thematic or geographic types, such as terrain, urban, or land-use maps (Zhou et al., 2018). These studies established the technical feasibility of map image analysis but largely emphasized semantic content rather than cartographic structure.

As data-driven methods matured, research began to extend toward generative and segmentation-based tasks. Paired datasets linking cartographic renderings with aerial or satellite imagery enabled image-to-image translation between different map representations, exemplified by the Pix2Pix Maps dataset (Isola et al., 2017). In parallel, segmentation-oriented datasets derived from USGS topographic maps or historical cartographic archives were developed to support automated extraction of roads, hydrography, and other geographic features (Saeedimoghaddam & Stepinski, 2020; Kramm et al., 2025). In parallel, large-scale historical map collections have also become available. The EPFL work on historical cartography assembled ADHOC

Records (771,561 catalog records aggregated from 38 institutions across 11 countries) and ADHOC Images (99,715 digitized map images), enabling computational investigations of cartographic evolution at an unprecedented scale (Petitpierre, 2025). VisTaxa developed a hierarchical taxonomy (51 taxa) by manually coding 400 historical visualization images and then scaling labels to the full OldVisOnline collection (13,511 images) via similarity-based and zero-shot CLIP-assisted prediction, offering a structured lens to quantify long-term trends in maps and other visualization forms (Zhang et al., 2025). While these efforts substantially advanced automated map digitization and feature extraction, they continued to conceptualize maps mainly as carriers of geographic information, with limited attention to design or layout characteristics.

More recent studies have increasingly recognized maps as structured visual documents composed of multiple cartographic elements arranged according to design conventions. This shift is reflected in the emergence of datasets that explicitly annotate map components such as titles, legends, scale bars, and inset maps. For example, CartoMark integrates map classification, text annotation, and style-related tasks within a unified benchmark (Zhou et al., 2024). The ubiMap and ubiMap-l datasets further provide element-level annotations across thousands of maps, enabling structural layout analysis and supporting learning-based models for layout retrieval and representation learning (Yang et al., 2025). Complementary to element- and layout-centric approaches, recent fragment-based studies have also explored the visual language of maps by decomposing map images into elementary graphical units and quantifying long-term stylistic evolution across large historical corpora. By analyzing changes in abstraction, graphical density, and visual composition, this line of work demonstrates how cartographic design and visual structure evolve over time beyond semantic content alone (Petitpierre et al., 2024). In addition to structural annotations, behavioral and perception-oriented datasets have also emerged. Eye-tracking datasets that record users' visual attention during map reading provide empirical evidence of how different map designs and layouts influence perception and cognitive processing (He et al., 2020). Such datasets complement image-based analyses by linking map structure to human interpretation, although they are typically limited in scale due to the cost of controlled experiments.

In summary, despite these advances, most existing map image datasets remain task-driven, with primary objectives centered on classification, segmentation, text detection, or retrieval. Cartographic design principles, particularly the color design and layout structure, and their large-scale empirical evaluation, are often addressed only implicitly or not at all. As a result, comprehensive, longitudinal analyses that directly examine how color and layout design characteristics evolve across large academic map corpora are still required.

# 3. Dataset construction

## 3.1 Journal selection and corpus scope

To construct a representative corpus for analyzing long-term map design practices, we selected a total of sixteen authoritative journals closely related to cartography and geographic information science, including eight Chinese-language journals and eight English-language journals. The journal information is shown in Table 1. For each selected journal, we collected all research articles published between 1990 and 2020. Although several journals were founded after 1990, they were still included in the corpus to ensure comprehensive coverage of the field within the selected time span. These include two Chinese journals—Journal of Geo-Information Science (founded in 1996) and Journal of Remote Sensing (founded in 1997)—and three English journals—Transactions in GIS (founded in 1997), Geoinformatica (founded in 1997), and Journal of Maps (founded in 2005). While publication records for most journals are continuous within the selected period, Transactions in GIS represents a notable exception: although founded in 1997, no articles were published in 1998, resulting in a one-year gap in its publication record.

The eight Chinese-language journals are fully open access and were downloaded directly from their official publishing platforms. English-language journals were accessed and downloaded through the Web of Science database. The number of articles collected from each journal is summarized in Table 1. In total, the corpus consists of 45732 articles across the sixteen journals. This corpus forms the basis for subsequent map extraction and design analysis.

Table 1. Overview of selected journals and article counts in the corpus

| Language | Journal name and its index in this study | Year founded | Article count |
|---|---|---|---|
| Chinese | Acta Geographica Sinica (地理学报) -A | 1934 | 3704 |
| | Science China Earth Sciences (中国科学：地球科学)-B | 1950 | 3633 |
| | Bulletin of Surveying and Mapping(测绘通报)-C | 1955 | 8798 |
| | Acta Geodetica et Cartographica Sinica (测绘学报)-D | 1957 | 3090 |
| | Geomatics and Information Science of Wuhan University(武汉大学学报(信息科学版)) -E | 1957 | 5101 |
| | Science of Surveying and Mapping(测绘科学)-F | 1976 | 4661 |
| | Journal of Geo-Information Science(地球信息科学学报)-G | 1996 | 1947 |
| | Journal of Remote Sensing(遥感学报) -H | 1997 | 2535 |
| | Total in Chinese | | 33469 |
| English | Annals of the American Association of Geographers-I | 1911 | 2516 |
| | Professional Geographer - J | 1949 | 2142 |
| | The Cartographic Journal - K | 1964 | 994 |
| | Cartography and Geographic Information Science - L | 1974 | 1063 |
| | International Journal of Geographical Information Science - M | 1987 | 2583 |
| | Transactions in GIS - N | 1997 | 1192 |
| | Geoinformatica - O | 1997 | 560 |
| | Journal of Maps - P | 2005 | 1213 |
| | Total in Chinese | | 12263 |
| | Total | | 45732 |

## 3.2 Automatic map extraction and map element identification

### 3.2.1 *Map extraction*

Maps in academic journals are distributed across heterogeneous PDF formats. While some articles provide digitally generated figures with explicit structural metadata, others are scanned documents in which maps are embedded as rasterized images, especially in some early publications. To robustly extract map images from both cases, we adopt a hybrid parsing and vision-based identification strategy. The process of map extraction is shown in Figure 2.

(1) PDF parsing and vision–language–based map identification

The maps from the article PDF files are identified with two stages: first, extracting images from PDF files, and second, identifying thematic maps from those images. Initially, we deploy MinerU-10B, a vision-based large model specialized in document recognition developed by OpenDataLab (Wang et al., 2024), to parse the PDF files and extract all constituent contents, including text blocks, tables, figures, and embedded images. This step ensures that we can recover candidate figure images regardless of whether the PDF is digitally authored or scanned. The associated caption text for each extracted image is also retained, which helps in further classification. Once the images are extracted, we proceed to identify thematic maps using a locally deployed Qwen 7B vision-language model, one of the most powerful open-source vision models developed by Alibaba (Wu et al., 2025). Specifically, a two-stage prompting strategy is employed. The first stage involves the model distinguishing thematic maps from other non-thematic figures, such as charts, diagrams, photographs, and schematic illustrations. This stage focuses on identifying figures primarily intended to represent geographic or spatial information. In the second stage, the model further filters out atypical or ambiguous cases that may remain after the initial classification, ensuring a more refined selection of maps. Through this process, we obtain 58,112 potential map images, though some of these figures visually resemble maps but are not suitable for design analysis, such as map-based statistical charts, heavily abstracted cartograms, or hybrid figures where geographic shapes are secondary to statistical graphics. Thus, a manual verification process is then performed.

(2) Manual verification

All automatically identified maps are subjected to manual verification to ensure their suitability for subsequent quantitative analysis. This manual screening serves two primary purposes: (1) retaining only maps with a clearly defined single dominant main map, and (2) excluding maps with ambiguous or unclear semantic content. The requirement of a single dominant main map arises from the methodological assumptions of the subsequent element detection and layout analyses, which depend on a clearly identifiable primary map region for consistent measurement and comparison.

For the first task, we filter out images with multiple main maps. If an image contains multiple subfigures and each subfigure can be treated as an independent and complete map, it is extracted as a separate map. However, if the subfigures cannot function independently, the entire image is excluded from further analysis. In the second task, we exclude maps that primarily serve as conceptual or schematic illustrations, such as principle-oriented diagrams or abstract illustrative maps. Additionally, maps whose semantic meaning or geographic reference is unclear are also discarded. This includes figures where geographic content is secondary or ambiguous, such as transportation diagrams or non-spatial visualizations that do not convey geographic or spatial information. After this manual identification and filtering process, 28,615 maps are retained for further analysis.

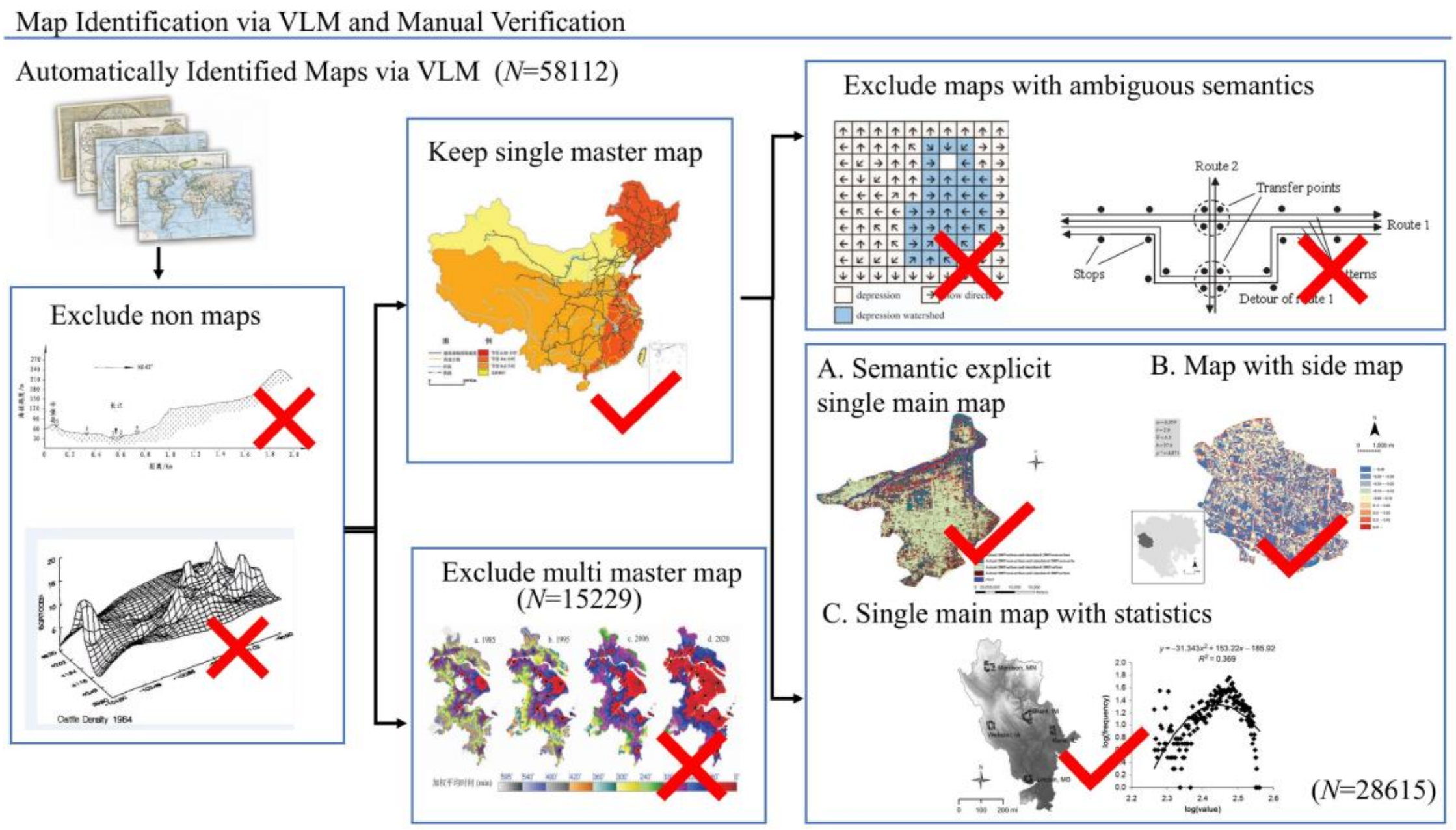

Figure 2. Workflow for extracting maps from article PDF files. A total of 58,112 potential maps are initially identified using a vision-language model (VLM) from the PDF versions of articles. Manual verification is then conducted to filter and refine the results, yielding 28,615 maps.

### 3.2.2 *Map element identification*

To enable fine-grained analysis of map design, we then identify the cartographic elements in the maps collected in ***Section* 3.2.1** and explicitly separate the main map from surrounding elements. This distinction supports two subsequent analyses in this study: layout analysis based on element arrangement and color analysis conducted exclusively on the main map. Accordingly, this process consists of three steps: map element categorization, map element identification, and main map extraction.

(1) Map element categorization

Based on established cartographic design literature (Tait, 2018; Jia et al., 2024), 10 common map elements that frequently appear in academic map layouts are adopted in this work. These elements include the main map, title, legend, scale bar, inset map, chart, descriptive text, north arrow, picture, and table. A detailed description is shown in Table 2.

Table 2. Types of map elements considered in this study and their definitions.

| **Element** | **Description** |
| --- | --- |
| Title | Textual element that provides the main subject or theme of the map, usually placed in a prominent position to guide the reader's attention. |
| Main map | The primary geographic area depicted, serving as the central focus of the layout. |
| Lengend | A reference box explaining the meaning of symbols, colors, or patterns used in the main map. |
| Scale bar | A graphical or numerical indicator that expresses the ratio between map distance and real-world distance. |
| Inset map | A supplementary map embedded within the layout, often used to show location context, enlargements, or global positioning. |
| Chart | The visualizations of statistical data, such as charts, graphs, or other forms of data representation designed to support the map's narrative. |
| Descriptive text | Explanatory or descriptive textual content (excluding the title), used to clarify background, context, or additional information. |
| North arrow | A directional indicator pointing to geographic north, assisting orientation. |
| Picture | Non-cartographic images (e.g., photos, diagrams) added to enrich the visual narrative of the map. |
| Table | Structured tabular information included within the layout, typically providing supporting statistical or categorical data. |

(2) Map element identification

To identify and localize map elements at scale, we adopt an annotation-and-detection pipeline similar to that used in visualization datasets such as VisImages (Zhang et al., 2025). A subset of representative maps (1,625 images) was manually annotated according to the taxonomy defined in Table 2. Each map was independently labeled by two annotators, and disagreements were resolved through discussion to reach consensus. The resulting annotations were used to fine-tune a YOLOv8-OBB object detection model, a state-of-the-art real-time detector supporting oriented bounding boxes, pretrained on a large-scale dataset (Mitrev et al., 2025). The trained model was evaluated on a held-out validation set and achieved stable detection performance. The key evaluation metrics reached satisfactory levels: mAP@0.5 stabilized around 0.83, and mAP@0.5:0.95 stabilized around 0.68. The model was then applied to the full corpus to automatically extract the positions and extents of map elements in a consistent manner.

During the element detection stage, 4,687 extracted map images were identified as lacking a recognizable main map and were therefore excluded from subsequent analysis, leaving 23,928 maps for further processing. These excluded cases mainly correspond to conceptual illustrations, schematic figures, or map-like graphics in which geographic content was not the dominant visual component. Since the subsequent analyses of color and layout design require a clearly defined main map as the structural and semantic anchor, these images were removed to ensure methodological consistency. An example of map element identification is shown in Figure 3(b).

(3) Main map extraction

Based on the oriented bounding boxes of the main map obtained during element identification, we further refine the main map regions using a segmentation-based approach. Specifically, the detected bounding box of the main map is used as a prompt for the Segment Anything Model (SAM), a foundation model for image segmentation with strong generalization capability across heterogeneous datasets (Kirillov et al., 2023), to generate a precise segmentation mask. Subsequent post-processing steps, including noise removal via Connected Component Analysis (CCA) (Rosenfeld & Pfaltz, 1966) is applied to retain only the primary geographic region while eliminating spurious fragments. The resulting main map masks serve as the basis for subsequent analyses of layout structure and color design, which focus exclusively on the map body. An example of the main map extraction result is shown in Figure 3(c).

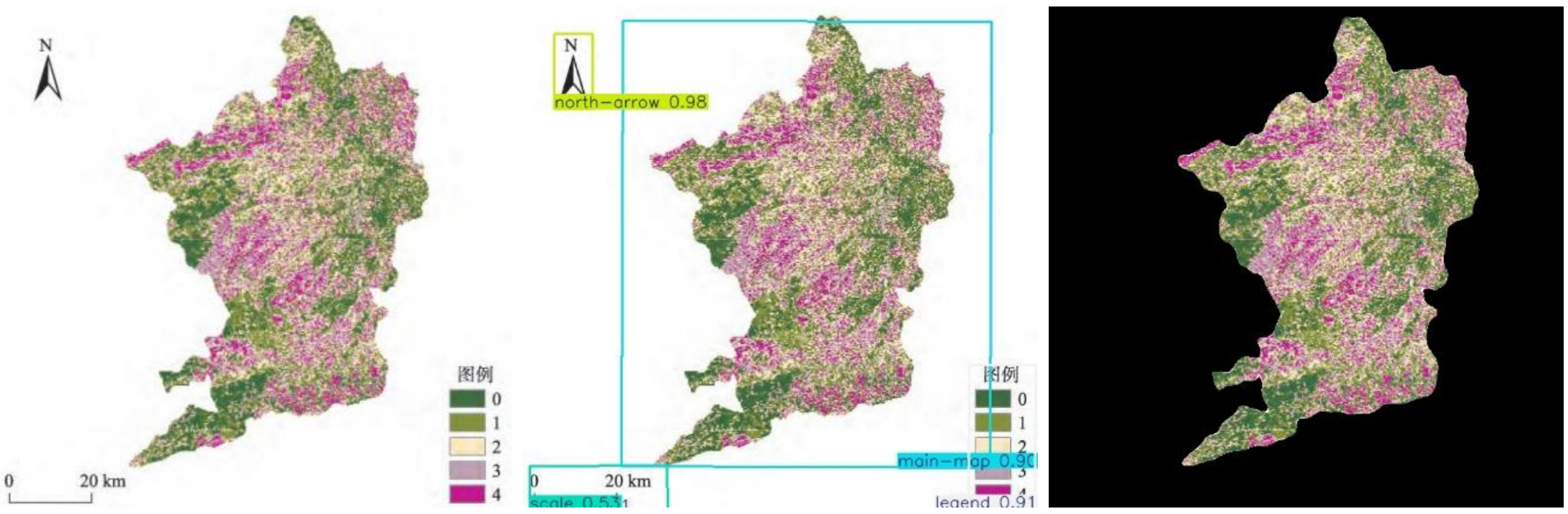

(a) (b) (c)

Figure 3. Workflow of map element identification and main map extraction. (a) Original map; (b) Detected map elements with oriented bounding boxes; (c) Extracted main map region after segmentation.

## 3.3 Dataset overview

As described above, 28,615 academic thematic maps are extracted from 45,732 research articles following the automatic extraction and manual verification procedures outlined in *Section* 3.1. Subsequently, based on the map element detection and main map extraction process introduced in *Section* 3.2, 4,687 maps lacking a recognizable dominant main map are removed, resulting in a final dataset of 23,928 maps for subsequent analysis. Table 3 summarizes the distribution of these 23,928 maps across journals, together with the corresponding article counts and map-to-article ratios ($N/C$).

As shown in Table 3, we can observe that Chinese-language journals contribute 15,971 maps, accounting for approximately two-thirds of the dataset, while English-language journals contribute 7,957 maps. This difference in absolute map counts largely reflects the underlying distribution of research articles, as Chinese-language journals collectively publish a substantially larger number of articles over the study period. When normalized by article count, we can reverse observe that English-language journals exhibit a higher average map density (*N*/*C* = 0.65) than Chinese-language journals (*N*/*C* = 0.48), indicating a stronger tendency to include cartographic figures per article in English-language publications. For the variation across individual journals, we can observe that journals with an explicit cartographic focus, such as The Cartographic Journal (*N*/*C* = 1.39) and Journal of Maps (*N*/*C* = 0.95), display the highest map densities, reflecting their emphasis on map-centered contributions. In contrast, more method- or application-oriented journals, such as Bulletin of Surveying and Mapping (*N*/*C* = 0.16) and Professional Geographer (*N*/*C* = 0.22), show considerably lower ratios, where maps typically play a supporting rather than central role. Several comprehensive GIS and remote sensing journals, including Acta Geographica Sinica, Science China Earth Sciences, and Transactions in GIS, exhibit map-to-article ratios close to or above one, suggesting that maps are routinely used as primary analytical artifacts in these venues.

Table 3. Distribution of academic maps across journals (1990–2020).

| Language | Journal Index | Number of maps (*N*) | Article count (*C*) | *N*/*C* |
|---|---|---|---|---|
| Chinese | A | 3820 | 3704 | 1.03 |
| | B | 4144 | 3633 | 1.14 |
| | C | 1397 | 8798 | 0.16 |
| | D | 718 | 3090 | 0.23 |
| | E | 1202 | 5101 | 0.24 |
| | F | 1651 | 4661 | 0.35 |
| | G | 1832 | 1947 | 0.94 |
| | H | 1207 | 2535 | 0.48 |
| | Total in Chinese | 15971 | 33469 | 0.48 |
| English | I | 1331 | 2516 | 0.53 |
| | J | 478 | 2142 | 0.22 |
| | K | 1385 | 994 | 1.39 |
| | L | 420 | 1063 | 0.40 |
| | M | 1687 | 2583 | 0.65 |
| | N | 1225 | 1192 | 1.03 |
| | O | 280 | 560 | 0.50 |
| | P | 1151 | 1213 | 0.95 |
| | Total in English | 7957 | 12263 | 0.65 |
| | Total | 23928 | 45732 | 0.52 |

The temporal distribution of maps is depicted in Figure 4. As shown in the figure, we can observe that there are distinct growth patterns between Chinese- and English-language academic journals. Both show an overall increase in the number of articles and maps published over time in Figure 4(a); however, the trends diverge in significant ways. In Chinese-language journals, there is a noticeable and substantial growth in both the number of maps and the number of published papers, particularly after the 2000s. This significant increase may be attributed to China's economic and technological boom in the 21st century, which led to the widespread adoption of digital tools, advancements in GIS and cartography, and a broader shift towards map-based data visualization in academic research. By contrast, English-language journals show a steadier increase, with relatively smaller growth in the number of maps published annually. This more gradual rise could be a result of English-language journals already having well-established publishing infrastructures and cartographic standards earlier on, limiting the scale of growth compared to Chinese journals, which experienced a period of catch-up after technological advancements became more accessible.

Regarding the map-to-paper ratio in Figure 4(b), English-language journals exhibit fluctuations in this ratio over time, but it remains relatively stable at a moderate level. This indicates that, while the number of maps increased, the ratio of maps to papers did not show dramatic changes, possibly due to the already established norms of academic publishing where maps are integral but not overrepresented. On the other hand, Chinese-language journals show a gradual increase in the map-to-paper ratio, reflecting a growing emphasis on maps in academic articles. This rise in the ratio could be due to the increasing importance of visual data representation in Chinese academic publishing, as well as the growing use of maps to convey complex spatial information in various fields of study.

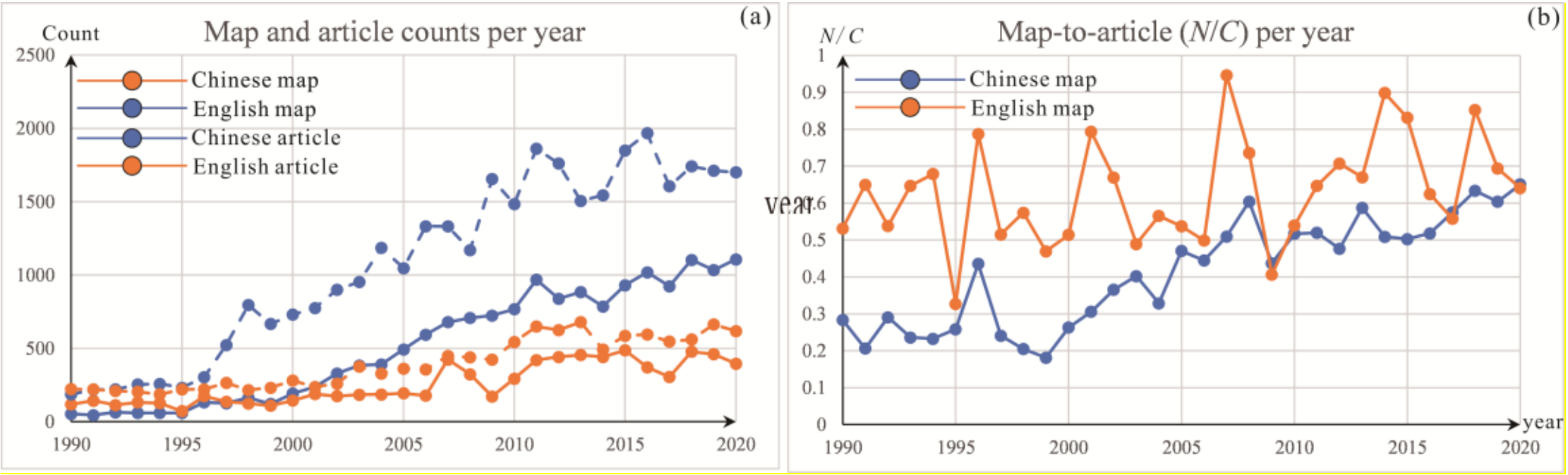

Figure 4. Cross-language temporal distribution of the academic articles and maps (1990–2020).

# 4. Design dimensions for analysis

To enable systematic analysis of map design practices, we characterize the collected maps from three complementary dimensions: map elements, color design, and layout structure. Map elements are identified and described in ***Section* 3.2**. Thus, this section mainly describes the definitions and metrics for color design (***Section* 4.1**) and layout structure (***Section* 4.2**), respectively.

## 4.1 Color design

Color design plays a central role in map interpretation by shaping visual tone, perceptual differentiation, and overall readability (Brewer, 2016). In this study, color analysis is conducted exclusively on the main map area, excluding legends and surrounding elements, to ensure that measured color properties reflect intrinsic map design rather than explanatory components. We characterize color design using two complementary principles: color tone and color complexity (Table 4). Color tone describes the intrinsic chromatic attributes of the map, including hue, saturation, and brightness, capturing the overall visual appearance of map colors. Color complexity characterizes the richness and organizational diversity of color usage, reflecting both the number of colors employed and their distribution across the map.

Table 4. Evaluation metrics for map color design.

| Principal | Indicator | Description and definition | Source |
| --- | --- | --- | --- |
| Color tone | $H_{main}$ | $H_{main}$ is the dominant hue of the map. | Brewer (2016) |
| | $S_{ave}$ | $S_{ave}$ is the average saturation value. | |
| | $B_{ave}$ | $B_{ave}$ is the average brightness value. | |
| Color complexity | $B_{con}$ | $B_{con}$ is the statistical variation of brightness values across the map, defined as $B_{con}=\sqrt{\frac{1}{N}\sum_{i=1}^{N}(B_i-B_{ave})^2}$ . $B_i$ represents the brightness value of the $i$-th pixel | Cao et al. (2025) |
| | $N_{hue}$ | $N_{hue}$ is the total number of distinct hues used in the main map, representing the basic richness of the color palette. To avoid counting marginal or noise colors, only hues accounting for at least 5% of the total pixel proportion in the main map are included in the calculation. | |
| | $E_{hue}$ | $E_{hue}$ is an entropy-based measure of hue distribution, and is defined as $E_{hue}=\sum_{i=1}^{N_{hue}} p(h_i)\log_2 p(h_i)$ , where $p(h_i)$ represents the portion of pixels of hue $h_i$ | |

## 4.2 Layout structure

Layout structure describes the spatial organization of the main map and auxiliary elements within a page (Tait, 2018). In this study, layout analysis focuses on the spatial arrangement of the main map and auxiliary elements detected on the map page. We characterize layout structure using four complementary principles: hierarchy, visual balance, compactness, and alignment (Table 5). Hierarchy describes the relative visual prominence of the main map compared with surrounding elements. Visual balance captures the spatial equilibrium of elements across the page. Compactness reflects the efficiency of space usage in arranging map components, while alignment characterizes the consistency of element positioning along shared axes.

Table 5. Evaluation metrics for map layout quality.

| Principal | Indicator | Description and definition | Source |
|---|---|---|---|
| Hierarchy | $D_{hier}$ | $D_{hier}$ measures the positional deviation of the main map by calculating the Euclidean distance between its centroid and the visual center of the page. Horizontal and vertical offsets are first normalized by page width and height, respectively, so that both $D_{horizontal}$ and $D_{vertical}$ fall within the interval [-1, 1]. The overall deviation ($D$) is computed as: $D = \sqrt{D^2_{horizontal} + D^2_{vertical}}$ After normalization, $D$ is rescaled to the range [0, 1], where 0 indicates perfect centering, and larger values represent increasing deviation from the visual center. | Dent et al. (2008) |
| Compactness | $R_{map}$ | $R_{map}$ measures the occupation of the main map relative to the total page and is defined as $R_{map}=A_{map}/A_{page}$, where $A_{map}$ and $A_{page}$ are the areas of the main map and the entire page, respectively. The ratio range is [0, 1], with larger values indicating better compactness. | Wei et al. (2017) |
| Alignment | $R_{horizontal}$ and $R_{vertical}$ | Alignment measures the structural consistency of layout elements and is defined as the proportion of misaligned sight lines relative to the total number of sight lines in the horizontal and vertical directions. $R_{horizontal} = \frac{C_{horizontal}}{L_{horizontal}}$, $R_{vertical} = \frac{C_{vertical}}{L_{vertical}}$ where $C_{horizontal}$ and $C_{vertical}$ are the number of misaligned horizontally and vertically sight lines, and $L_{horizontal}$ and $C_{vertical}$ are the total number of sight lines in the corresponding directions. The ratio range | O'Donovan et al. (2014) |
| | | is [0, 1], with smaller values indicating better alignment. | |
| Visual balance | $B_{horizontal}$ and $B_{vertical}$ | Visual balance is evaluated by using weighted contributions of map elements, considering both horizontal (top-bottom) and vertical (left-right) balance to ensure layout stability. The balance index is defined as the normalized difference between the visual weights on each side: $$B = \frac{W_1 - W_2}{W_1 + W_2}$$ Where $W_1$ and $W_2$ represent the total visual weights of elements on the two opposing sides (top-bottom and left-right). Thus, it has two visual balances in top-bottom ($B_{horizontal}$) and left-right ($B_{vertical}$), and falls within [-1, 1], 0 indicates perfect equilibrium. | Wei et al. (2017) |

# 5. Statistical analysis and results

## 5.1 Overall differences between Chinese and English maps

### 5.1.1 Map element

The cross-language differences in map elements are analyzed from three perspectives: (1) element quantity per map, (2) element type usage frequency, and (3) element co-occurrence patterns. To account for differences in total map counts between Chinese and English-language journals, all analyses are conducted on a proportional basis to ensure comparability. The analysis results are shown in Figure 5 and Table 6.

**In terms of the number of elements per map**, Figure 5(a) shows that Chinese and English academic maps exhibit highly similar overall distributions. In both groups, maps with a small number of elements dominate. Specifically, maps containing one to three elements account for 79.3% of Chinese maps and 75.2% of English maps. The single-element and two-element cases alone represent 55.6% of Chinese maps and 54.4% of English maps, indicating that concise layouts are the prevailing design choice in both language contexts. When summarized at the aggregate level, English-language maps contain a slightly higher average number of elements per map than Chinese-language maps (English: 2.59 elements per map; Chinese: 2.53 elements per map). While English maps show a slightly higher proportion of maps with four or more elements (approximately 24.8% compared to 20.7% in Chinese maps), this difference appears mainly in the long tail of the distribution and does not alter the overall similarity of element-count patterns between the two groups.

**In terms of the element type usage**, Figure 5(b) shows that the relative frequencies of most cartographic elements are remarkably consistent across Chinese and English journals. For example, legends and scale bars are the most frequently used auxiliary elements in both groups. Legends appear in 45.48% of Chinese maps and 50.02% of English maps (much higher), while scale bars appear in 30.33% and 33.09%, respectively. North arrows show almost identical usage rates (16.72% in Chinese maps and 16.55% in English maps). Other elements, such as titles, inset maps, pictures, tables, and charts, remain relatively infrequent in both datasets, with usage rates generally below 20%. Although English maps exhibit slightly higher frequencies for descriptive text (3.27% vs. 1.72%) and tables (0.60% vs. 0.43%), these elements account for only a small fraction of the overall layouts and thus represent localized rather than structural differences.

**In terms of element co-occurrence patterns**, it is implemented via the Apriori algorithm (Huang et al., 2000), and the results are shown in Table 6. Table 6 further confirms the strong similarity in compositional conventions. In both language groups, the most common element combination is the pairing of the main map with a legend, occurring in 62.29% of Chinese maps and 70.60% of English maps. The inclusion of a scale bar together with the main map is also common in both groups (41.54% in Chinese maps and 46.71% in English maps). Higher-order combinations, such as the joint presence of legend, scale bar, and north arrow, follow the same rank ordering across languages, although their absolute frequencies are moderately higher in English maps. These differences indicate that English maps tend to adopt slightly more fully specified combinations, but they do not alter the dominant co-occurrence structure shared by both groups.

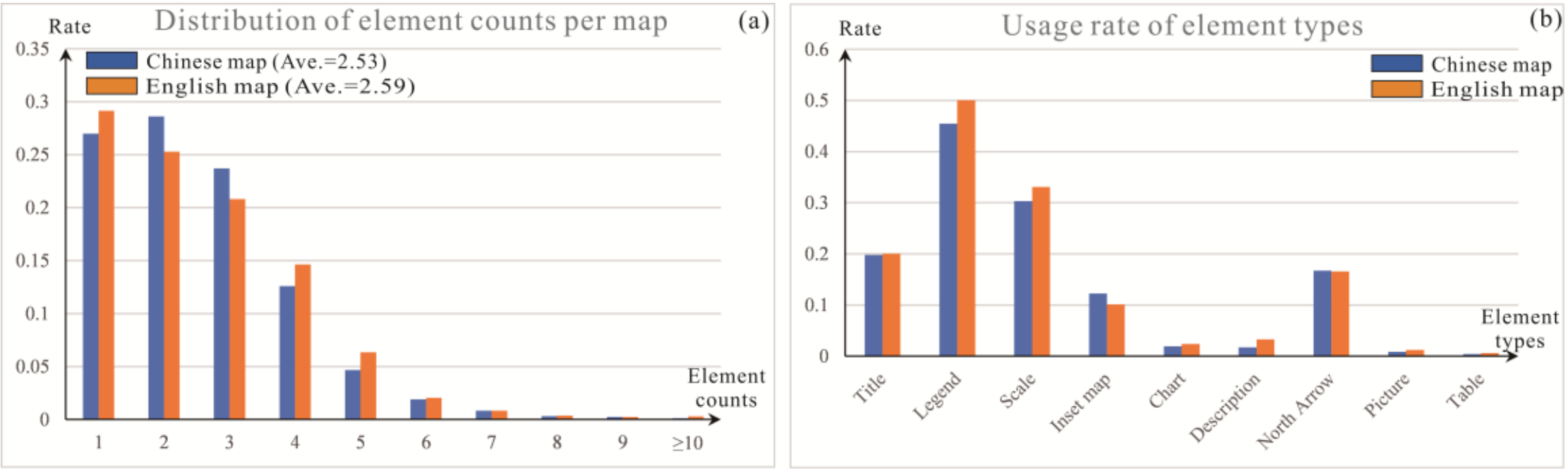

Figure 5. Statistics of detected cartographic elements in the dataset.

Table 6. Co-occurrence of map elements in Chinese- and English-language academic maps.

| Cooccurrence Type | Chinese | | English | |
|---|---|---|---|---|
| | **Element Combination** | **Rate (%)** | **Element Combination** | **Rate (%)** |
| Top-2 | Legend<br>Main map | 62.29% | Legend<br>Main map | 70.60% |
| | Scale bar<br>Main map | 41.54% | Scale bar<br>Main map | 46.71% |
| | Map title<br>Main map | 27.14% | Legend<br>Scale bar | 31.86% |
| Top-3 | Scale bar<br>Legend<br>Main map | 23.33% | Scale bar<br>Legend<br>Main map | 31.86% |
| | Legend<br>Main map<br>North arrow | 13.51% | Legend<br>Main map<br>North arrow | 17.83% |
| | Map title<br>Legend<br>Main map | 12.94% | Map title<br>Legend<br>Main map | 17.14% |

### 5.1.2 Color design

The cross-language differences in color design are analyzed based on the six indicators in Table 4. All color features are extracted using OpenCV in the HSB color space to ensure computational consistency. Hue values are classified according to the OpenCV hue scale and grouped into ten discrete categories: black, gray, white, red, orange, yellow, green, cyan, blue, and purple (Cao et al., 2025). Furthermore, we conduct Mann-Whitney U tests for the five quantitative indicators to assess cross-language differences, including $S_{ave}$, $B_{ave}$, $B_{con}$, $N_{hue}$, and $E_{hue}$. The statistical results, together with the distributional comparisons, are presented in Figure 6.

**(1) Color tone**

Figure 6(a) shows that white is the main dominant hue in both language groups, accounting for approximately 43% of Chinese maps and 40% of English maps. Gray tones account for roughly 25% in both datasets. Chromatic hues such as red, orange, yellow, green, and blue each individually account for less than 10%, indicating that academic maps in both contexts rely primarily on neutral or low-saturation dominant hues.

For average saturation ($S_{ave}$) in Figure 6(c), English-language maps exhibit significantly higher values than Chinese-language maps ($p$ = 0.000***). The median saturation for Chinese maps is 0.020, whereas the median for English maps is 0.075, indicating that English maps tend to employ relatively more saturated color schemes. However, despite this statistically significant difference, the overall saturation levels in both language groups remain low. In both datasets, approximately 75% of maps have average saturation values below 0.25, suggesting that low-saturation or muted palettes dominate academic cartographic design. At the same time, the upper tails of the distributions indicate the presence of a small number of maps with markedly higher saturation values, reflecting occasional adoption of more vivid or bold color schemes. These results suggest that while English-language maps show a stronger tendency toward chromatic expressiveness, both language groups largely adhere to restrained and academically conservative tonal conventions.

For average brightness ($B_{ave}$) in Figure 6(c), Chinese-language maps display slightly higher values than English-language maps. The median brightness is 0.843 for Chinese maps and 0.836 for English maps. Although the statistical test indicates that the difference is significant ($p$ = 0.005**), the magnitude of the difference remains small, reflecting only a subtle tonal variation between the two groups. Importantly, both language groups exhibit generally high brightness levels, with approximately 75% of maps concentrated in the upper brightness range, reflecting a dominant preference for light tonal compositions. This suggests that academic maps in both contexts commonly adopt bright backgrounds and high overall luminance, likely to ensure legibility and printing clarity, with only minor cross-language distinctions.

**(2) Color complexity**

Figure 6(b) reveals a statistically significant difference in hue richness ($N_{hue}$) between the two language groups ($p$ = 0.000***). While maps using two hues dominate both groups (44.8% in Chinese and 38.1% in English), English-language maps show higher proportions in the three-hue (27.2% vs. 21.6%) and four-hue categories (13.2% vs. 11.2%). English maps also exhibit slightly longer tails in higher hue-count categories, indicating greater chromatic diversity. However, despite these differences, the overall distribution remains strongly concentrated in low hue-count categories. In both language groups, maps using no more than four hues account for over 85% of the total, demonstrating a shared tendency to avoid excessive chromatic complexity. This suggests that academic cartographic design in both contexts favors restrained palette structures, maintaining controlled hue variation while allowing limited flexibility in chromatic richness.

For brightness contrast ($B_{con}$) in Figure 6(c), no statistically significant difference is observed between the two groups ($p = 0.761$). The median values are nearly identical, at 0.169 for Chinese maps and 0.170 for English maps, and the violin distributions largely overlap. Furthermore, the distributions show strong overlap, with almost all maps exhibiting brightness contrast values below 0.4, and approximately 75% of maps concentrated below 0.2 in both datasets. These results indicate that academic maps in both language contexts consistently employ low levels of tonal contrast, avoiding strong luminance differentiation across spatial units. This pattern suggests a shared design preference for moderate brightness variation, likely to maintain visual coherence and reduce perceptual distraction rather than to emphasize dramatic tonal segmentation.

For color entropy ($E_{hue}$) in Figure 6(d), a statistically significant cross-language difference is observed ($p = 0.000$***), with English-language maps exhibiting higher values overall. The mean entropy is 1.270 for English maps and 1.155 for Chinese maps, indicating relatively greater chromatic diversity in English-language publications. However, the overall distribution remains concentrated in lower entropy ranges in both groups. Approximately 75% of maps exhibit entropy values below 1.8, suggesting generally limited chromatic dispersion. This finding is consistent with the $N_{hue}$ results, where the majority of maps employ no more than four distinct hues. In addition, a noticeable proportion of maps exhibit entropy values close to zero. This corresponds to the presence of single-hue maps, which account for about 10% in both Chinese and English datasets (as shown in Figure 6(b)). Taken together, while English-language maps display moderately higher color entropy, both language groups predominantly adopt restrained color distributions, reinforcing the overall pattern of controlled palette usage observed in the hue-count analysis.

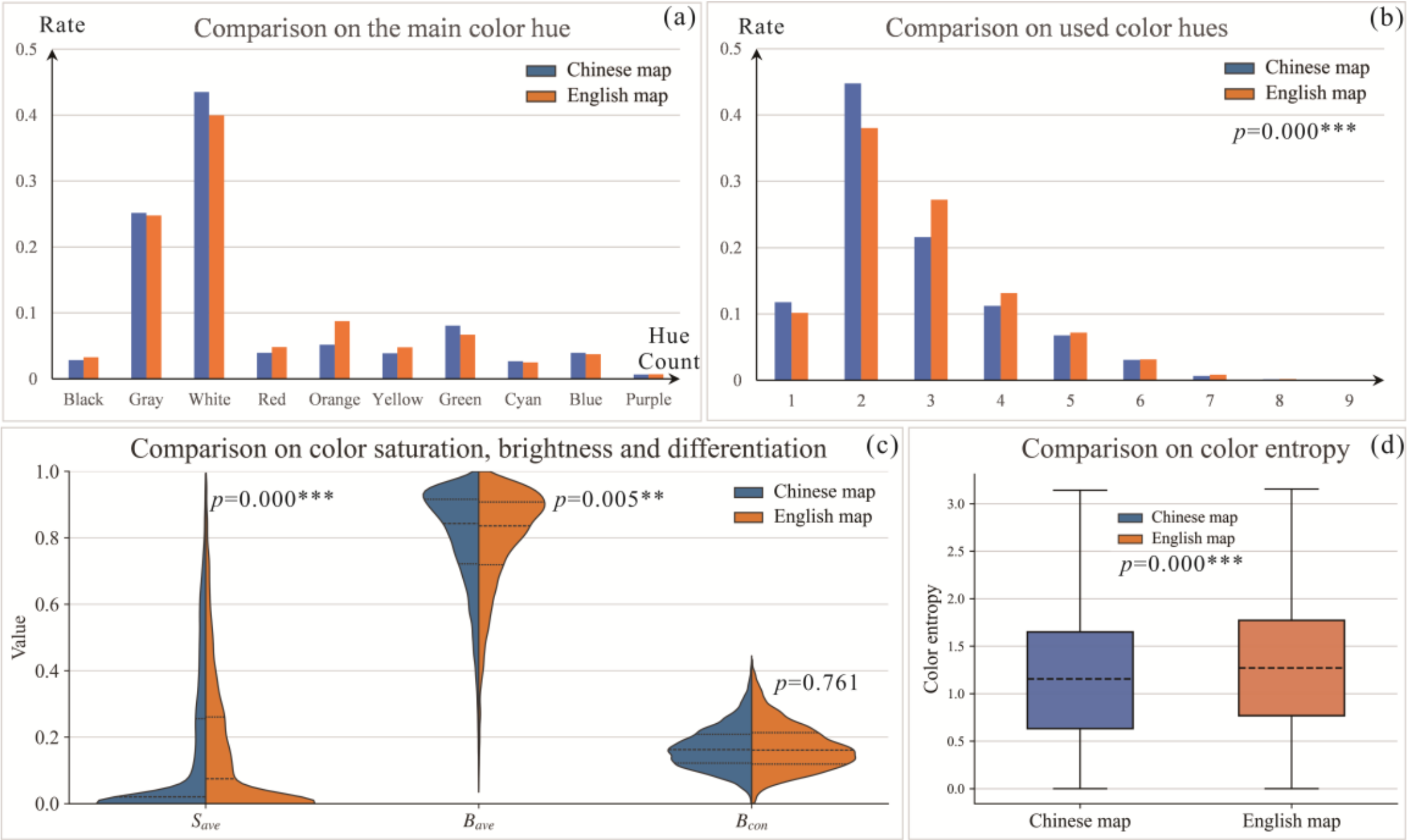

Figure 6. Cross-language comparison of color design indicators between Chinese- and English-language academic maps.

### 5.1.3 Layout structure

The cross-language differences in layout structure are analyzed based on the six indicators in Table 5. The Mann-Whitney U tests are conducted to examine cross-language differences for each quantitative indicator. The results are shown in Figure 7. As shown in the figure, the overall layout structures of Chinese and English academic maps are highly comparable. Although several indicators reach statistical significance, the magnitudes of the differences are consistently small, and the distributions exhibit substantial overlap.

Specifically, for hierarchy ($D_{\text{hier}}$) in Figure 7(a), its mean value is 0.101 for Chinese maps and 0.108 for English maps, indicating nearly identical centering practices. The statistical test confirms that the difference is not significant ($p$ = 0.926). In both language groups, the distributions are concentrated in the lower deviation interval (below 0.2), demonstrating that academic maps consistently prioritize centered composition.

For compactness ($R_{\text{map}}$) in Figure 7(a), a statistically significant difference is observed ($p$ = 0.000***), yet the magnitude of the difference remains limited. The mean value is 0.765 for Chinese maps and 0.781 for English maps, corresponding to a gap of only 0.016. Importantly, the distributions reveal strong concentration in higher compactness ranges. In both language groups, nearly all maps exhibit $R_{\text{map}}$ values above 0.5, and approximately 75% of maps exceed 0.7, indicating that the main map typically occupies a large proportion of the page. Although English-language maps show slightly higher compactness on average, the substantial distributional overlap suggests that both groups adopt a similarly main-map-dominant layout structure.

Alignment in Figure 7(a), inconsistency remains comparable across languages. For horizontal alignment ($R_{\text{horizontal}}$), the mean is 0.596 (Chinese) versus 0.580 (English) ($p = 0.008^{**}$). For vertical alignment ($R_{\text{vertical}}$), the mean is 0.641 (Chinese) versus 0.595 (English) ($p = 0.005^{**}$). While these differences are statistically detectable, the absolute gaps (0.016 and 0.046, respectively) remain small relative to the full range [0,1], and the interquartile ranges substantially overlap, indicating only modest cross-language variation. One noteworthy finding, however, diverges from conventional expectations. Alignment is typically emphasized as a fundamental cartographic principle, implying that well-designed maps should exhibit strong structural consistency. Yet the mean values in both language groups are close to 0.6, suggesting a moderate level of alignment inconsistency rather than strict geometric regularity. Moreover, the distributions span a wide range, with a considerable number of maps exhibiting values close to 0 (indicating high alignment quality), alongside many maps with relatively high inconsistency values. This dispersion suggests that alignment is not uniformly enforced across academic maps. Instead, it appears to function as a flexible design dimension rather than a rigid rule. While a subset of maps adheres closely to formal alignment principles, others adopt more relaxed or adaptive strategies, possibly balancing structural order with content-driven layout needs.

The differences are minimal in visual balance, as shown in Figure 7(b). For horizontal visual balance ($B_{\text{horizontal}}$), the mean value is 0.286 (Chinese) and 0.289 (English) ($p = 0.111$), indicating no statistically significant difference. For the vertical balance ($B_{\text{vertical}}$), the mean value is 0.182 (Chinese) and 0.184 (English) ($p = 0.000^{***}$). Although vertical balance reaches statistical significance, the difference in mean values is less than 0.002, which is also practically negligible. Across both language groups, the balance indices are predominantly positive, with mean values of approximately 0.28 for $B_{\text{horizontal}}$ and 0.18 for $B_{\text{vertical}}$. These values indicate a modest but consistent compositional tendency—slightly upward in vertical placement and mildly asymmetric in horizontal distribution. Nevertheless, all values remain relatively close to zero, indicating that despite these subtle tendencies, academic maps in both language contexts continue to prioritize overall equilibrium and balanced composition.

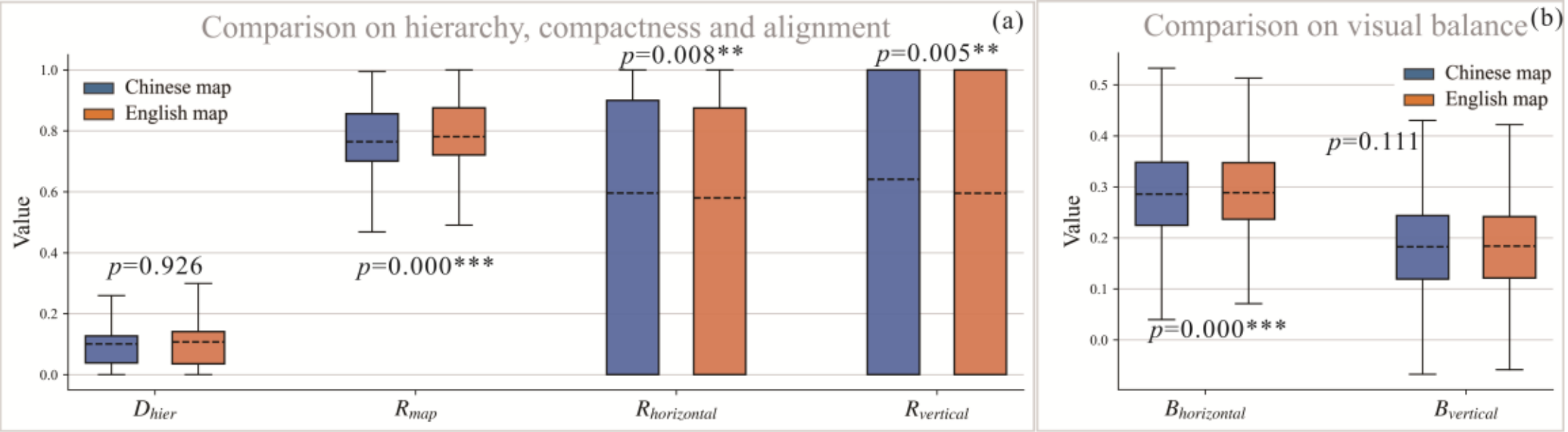

Figure 7. Cross-language comparison of layout structure indicators between Chinese- and English-language academic maps.

## 5.2 Temporal evolution within each language

### 5.2.1 Map element

To analyze the temporal evolution within each language for map elements, we examine (1) the average number of elements per map and (2) the presence ratios of key auxiliary elements (legend, scale, and north arrow that are identified in *Section* 5.1.1). To quantitatively assess whether these variables exhibit monotonic trends over time, Spearman's rank correlation tests are conducted between publication year and each element-related indicator. The results are shown in Figure 8.

The results in Figure 8 show a clear and statistically supported increasing trend in map element richness over time. The average number of elements per map in Figure 8(a) increases from approximately 1.83 (Chinese) and 2.08 (English) in 1990 to around 3.01 and 3.03 in 2020, respectively. Spearman analysis confirms the significant positive correlation between year and element count for both language groups ( $\rho = 0.875, p = 0.000***$ for Chinese maps and $\rho = 0.813, p = 0.000***$ for English maps), indicating a consistent long-term upward trend. At the element level in Figures 8(b), (c) and (d), legend usage exhibits the strongest growth, rising steadily throughout the study period and reaching about 60% by 2020. The scale bar and north arrow usage also show a more moderate but clear increase. But their usage remains comparatively low (not exceed 40%). Together, these results indicate that academic thematic maps have progressively converged toward a relatively stable configuration—typically comprising around three elements per map, with legends present in a majority of cases (60%)—reflecting a growing emphasis on informational completeness rather than purely minimalist design.

Although Chinese- and English-language journals share the same overall increasing trends, we can also observe that their temporal patterns differ both at the aggregate level and at the element level. As shown in Figure 8(a), in the early period, English-language maps seem to contain more elements on average than Chinese-language maps (e.g., 2.08 vs. 1.83 in 1990). Chinese-language journals exhibit a faster increase after the mid-2000s, leading to convergence by 2020 (both around 3.0). However, this advantage is not uniform across element types. Similar convergence patterns are observed in the presence ratios of legends and scale bars, as shown in Figures 8(b) and (c). Spearman correlations further indicate that the strength of the temporal association is comparable across the two language groups, suggesting that earlier cross-linguistic differences diminish rather than reverse over time. In contrast, north arrows in Figure 8(d) constitute a notable exception. English-language journals exhibit an earlier and more pronounced increase in north arrow usage during the late 1990s and early 2000s, leading to higher presence rates throughout much of the 2000s. In contrast, Chinese-language journals remain at relatively low levels until the mid-2000s (with most years below 5%), after which north arrow usage increases steadily and eventually surpasses that of Chinese-language maps in the later years.

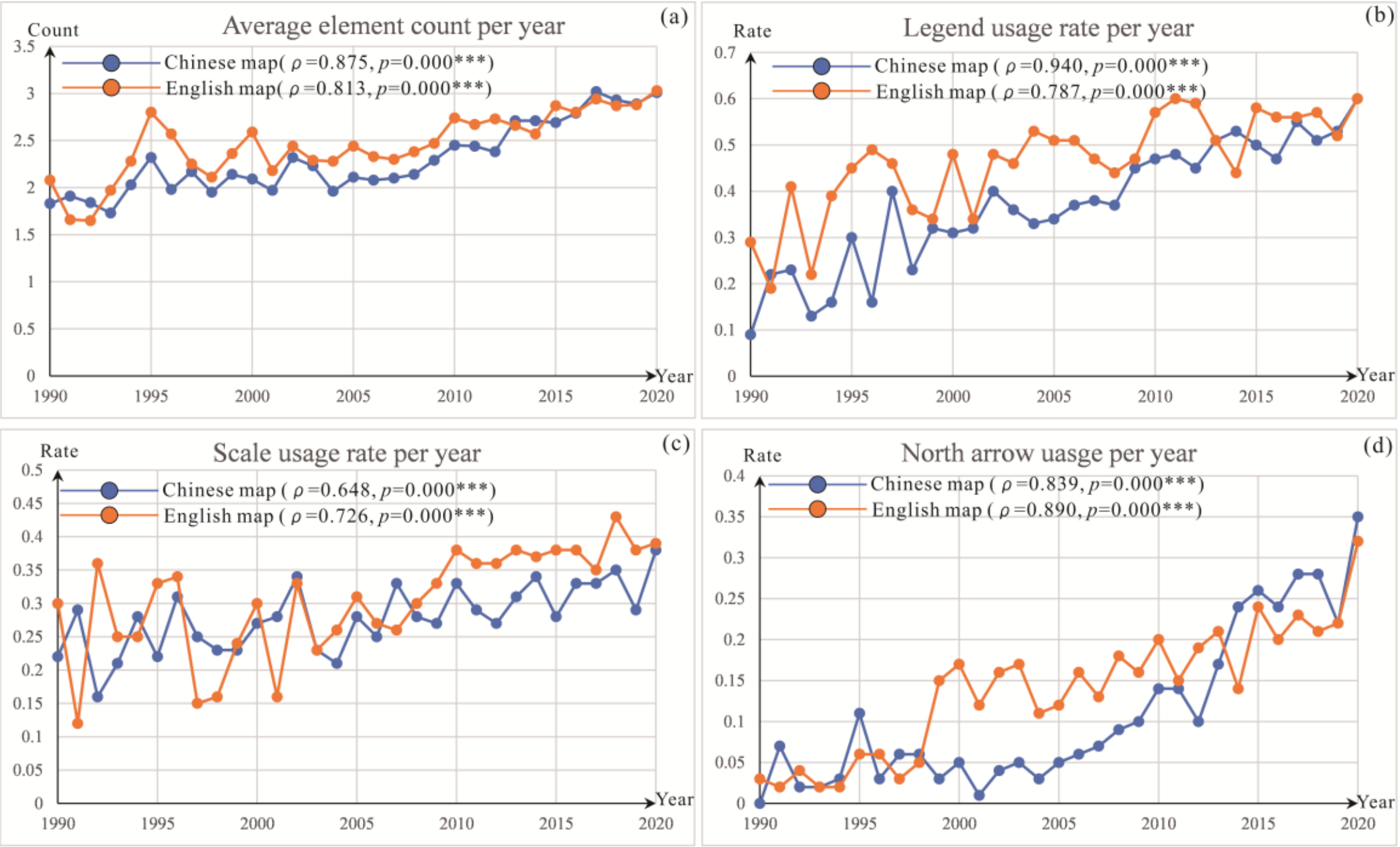

Figure 8. Temporal evolution of map element configuration in Chinese- and English-language academic maps (1990–2020).

### 5.2.2 Color design

To analyze the temporal evolution of color design within each language, we examine the annual mean values of six indicators in Table 4. As shown in the analysis in *Section* 5.1.1, black, white, and gray account for the majority of dominant hues in both language groups. Therefore, in analyzing the dominant hue in maps over time, we focus on the proportion of chromatic hues (i.e., excluding black, white, and gray) to assess whether the use of colored dominant tones changes across years. To quantitatively assess monotonic temporal trends, Spearman's rank correlation tests are conducted between publication year and each quantitative indicator.

The results are illustrated in Figure 9. As shown in the figure, we can observe that the two language groups exhibit largely parallel temporal trajectories across most color indicators, with consistent directions of change over the thirty-year period. First, several indicators exhibit clear increasing trends over time. The usage rate of chromatic dominant hues in Figure 9 (a) rises substantially in both language groups (Chinese: $\rho = 0.922$; English: $\rho = 0.934$, $p$=0.000***), increasing from near-zero levels in the early 1990s to approximately 0.45 by 2020. Similarly, average saturation ($S_{ave}$) in Figure 9(b) shows strong positive correlations with year (Chinese: $\rho = 0.921$; English: $\rho = 0.947$, $p$ = 0.000***), rising from below 0.05 to above 0.20. Color complexity indicators in Figures 9(e) and (f) follow the same pattern: the number of used hues ($N_{hue}$) increases from roughly 2.1-2.3 to about 3.0 (Chinese: $\rho = 0.841$; English: $\rho = 0.867$, $p$ = 0.000***), and color entropy ($E_{hue}$) grows from approximately 0.8-0.9 to around 1.4-1.5 (Chinese: $\rho = 0.931$; English: $\rho = 0.892$, $p$ = 0.000***). Together, these increases indicate a gradual enrichment of chromatic diversity and greater adoption of color in academic cartography, possibly reflecting technological advances in digital production and evolving aesthetic standards.

Second, brightness contrast ($B_{con}$) in Figure 9(d) shows a consistent decreasing trend in both groups (Chinese: $\rho = -0.807$; English: $\rho = -0.939$, $p$ = 0.000***), declining from approximately 0.22–0.24 to around 0.15 by 2020. This suggests that while hue diversity increases, tonal differentiation becomes more moderate over time, pointing toward smoother visual transitions and enhanced compositional coherence.

Third, average brightness ($B_{ave}$) in Figure 9(c) remains largely stable. English-language maps show no significant temporal trend ( $\rho = -0.265$ , p = 0.150), while Chinese-language maps display a slight decline ( $\rho = -0.907$ , p = 0.000***). However, both groups consistently remain within a high-brightness range (approximately 0.75 – 0.85), indicating sustained preference for light tonal compositions.

Although the overall evolutionary directions are broadly similar, some annuanced differences can be observed. For example, while both groups show increasing trends in $N_{hue}$ and $E_{hue}$, the magnitude of growth is slightly more pronounced in English-language maps, resulting in consistently higher entropy values in the later years. Regarding color tone, $B_{ave}$ in Chinese-language maps shows a small but noticeable decline, whereas English-language maps remain largely unchanged. However, the decline in Chinese maps is gradual and limited in scale, and by the end of the study period, the brightness levels of the two groups remain close. For saturation, both groups exhibit slight upward trends, but English-language maps consistently maintain higher saturation values across the entire period. These patterns suggest parallel evolution with persistent but moderate stylistic differences.

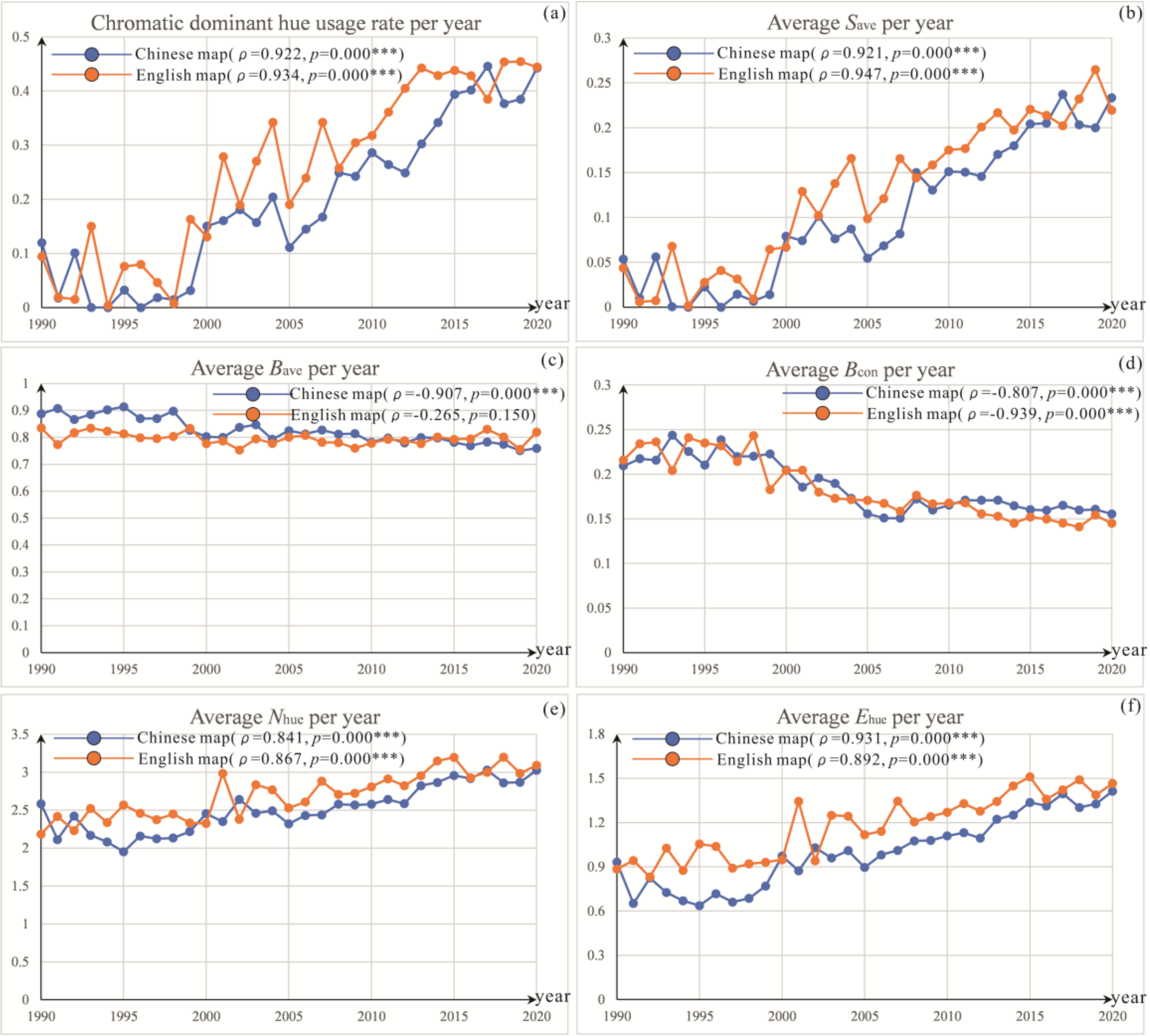

Figure 9. Temporal evolution of color design indicators in Chinese- and English-language academic maps (1990–2020).

### 5.2.3 Layout structure

To examine the temporal evolution of layout structure within each language group, we analyze the annual mean values of the layout indicators in Table 5. Spearman’s rank correlation tests are conducted between publication year and each indicator to evaluate the presence of monotonic temporal trends. The results are illustrated in Figure 10.

From the figure, we can observe that the Chinese- and English-language maps exhibit broadly similar and relatively stable temporal patterns across most layout indicators. Among all indicators, compactness ($R_{map}$) in Figure 10(b) is the only indicator that demonstrates a consistent decreasing trend in both language groups. Specifically, $R_{map}$ declines from approximately 0.82 in the early 1990s to around 0.75 by 2020, indicating that the proportion of page area occupied by the main map has gradually reduced over time. This suggests that auxiliary elements increasingly compete for layout space, reflecting a gradual shift toward more information-dense compositions. In contrast, several structural indicators show no meaningful temporal change. Visual balance indices ($B_{\text{horizontal}}$ and $B_{\text{vertical}}$) in Figures 10(e) and (f) likewise exhibit no significant trends in either language group, fluctuating within narrow intervals without systematic drift. These results suggest strong persistence of equilibrium-based layout conventions. A partial divergence appears in certain indicators where statistical significance is detected in Chinese maps but not in English maps. For example, $D_{\text{hier}}$ in Chinese maps shows a significant positive correlation with year ($\rho = 0.809$, $p$ = 0.000***), while English maps show no trend. Similarly, $R_{\text{vertical}}$ increases significantly in Chinese maps ($\rho = 0.842$, $p$ = 0.000***), whereas English maps remain stable ($\rho = 0.250$, $p$ = 0.175). However, closer examination reveals that these statistically detectable trends correspond to relatively small changes in magnitude. $D_{\text{hier}}$ values remain largely confined between 0.08 and 0.12, and $R_{\text{vertical}}$ shifts only moderately within the mid-range of the scale. Therefore, although temporal drift is detectable in Chinese maps, the practical effect on the overall layout structure is limited.

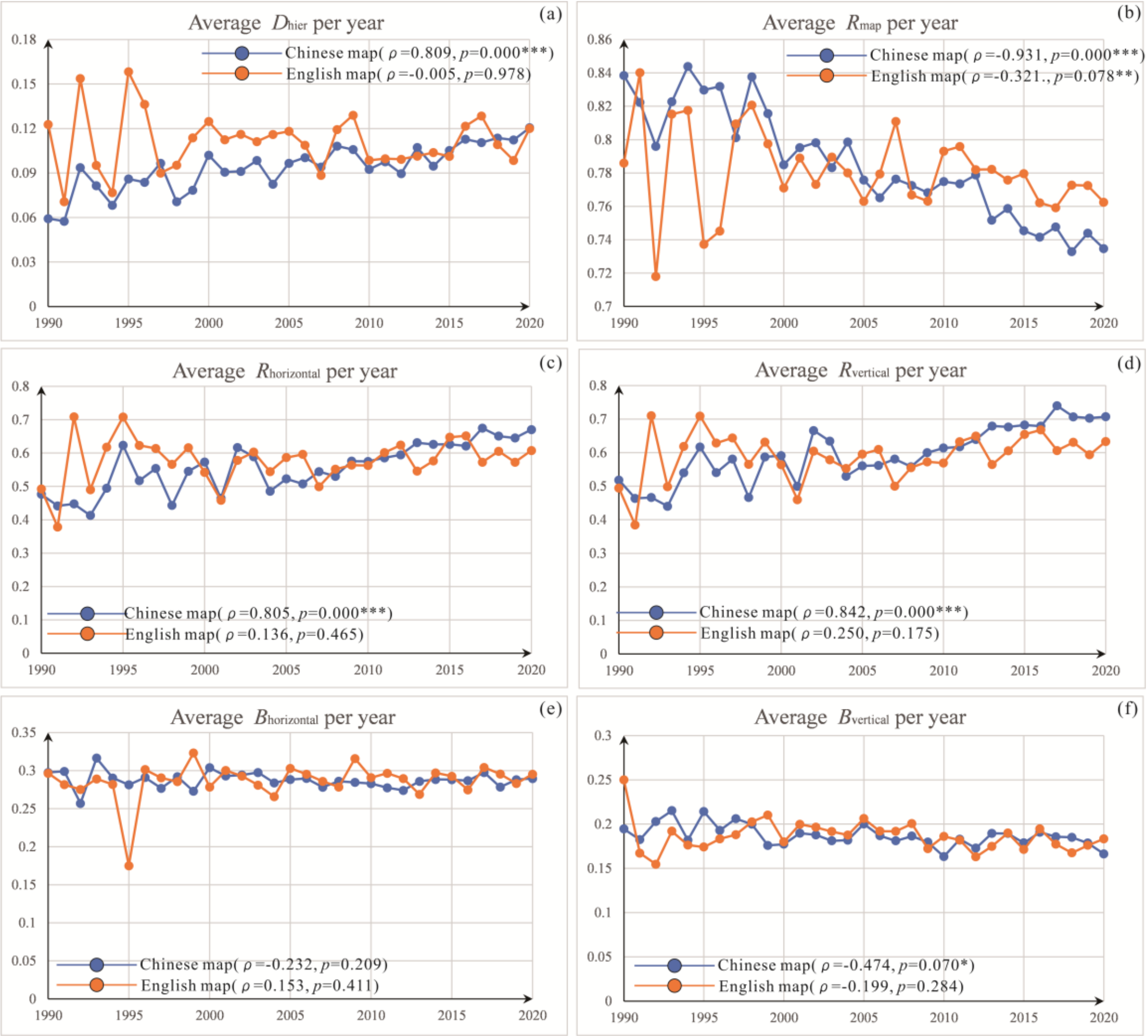

Figure 10. Temporal evolution of layout structure indicators in Chinese- and English-language maps (1990－2020).

# 6. Discussion

## 6.1 Cultural and academic influences on map design practices

Maps are cultural and academic artifacts, and their design may be inherently influenced by cultural and academic factors, which shape their elements, such as color schemes, layout structures, and the use of map components. Our study reveals these influences, highlighting the distinct cultural preferences and academic traditions that impact cartographic design.

**(1) Cultural influences on map design**

For color design, Chinese-language maps demonstrate a stronger reliance on muted and neutral hues. As shown in Figure 6, 68.8% Chinese maps are dominated by gray or white tones and only 14.8% employing more than four hues, reflecting the traditional cultural preference for subtlety and clarity. In contrast, English-language maps show relatively greater chromatic diversity, with 64.8% dominated by gray or white tones and 29.1% using more than four hues, aligning with Western cartographic traditions that emphasize clear distinctions and visual appeal. This difference can be traced back to historical cultural norms that influenced visual expression, such as the use of color in traditional Chinese art, which often emphasizes harmony and modesty, as opposed to the Western focus on contrast and differentiation.

Moreover, layout practices in Chinese and English maps can be shaped by differing expectations of information presentation and have been observed to show significant differences in compactness ($R_{\text{map}}$), as shown in Figure 7. Chinese academic maps, for instance, may often prioritize the integration of multiple elements such as legends, scale bars, and descriptive texts within the map frame, reflecting a more information-dense and content-rich design. In contrast, English-language maps tend to focus on the primary geographic area, with auxiliary elements placed more sparingly, allowing the map itself to serve as the central informational source. These differences highlight the diverse expectations of map function and form across cultures, with Chinese maps favoring informative clarity and English-language maps prioritizing visual coherence.

**(2) Academic influences on map design**

Beyond cultural preferences, the temporal patterns observed in *Section* 5 suggest a strong institutional and academic influence on map design practices. As shown in Figure 8, noticeable cross-language differences existed in the early 1990s. In the early 1990s, English-language maps exhibited higher structural completeness, with average element counts of 2.08 compared to 1.83 in Chinese-language maps. Legend usage also showed a marked gap (0.29 vs. 0.09), as did scale bar usage (0.30 vs. 0.22). However, by 2020, both groups converged to similar configurations: approximately three elements per map, legend usage near 0.6, and scale usage around 0.39. A similar convergence is evident in color and layout dimensions, as shown in Figures 9 and 10. Hue richness ($N_{\text{hue}}$), initially different (2.58 in Chinese vs. 2.18 in English maps), increased in both groups and stabilized near 3.05. Hierarchical deviation ($D_{\text{hier}}$) and vertical balance ($B_{\text{vertical}}$), which also exhibited early-stage differences, ultimately aligned at comparable levels ($D_{hier}$ ≈ 0.12; $B_{vertical}$ ≈ 0.17). These synchronized trajectories across multiple indicators indicate that early design differences were not persistent structural divergences but transitional stages. English-language journals, with longer-established international publication traditions and standardized figure guidelines, may have institutionalized certain cartographic conventions earlier. Chinese-language journals, particularly after the widespread adoption of digital production workflows and increasing internationalization after the 2000s, progressively aligned with similar standards. Thus, the temporal evidence indicates that early differences likely reflected localized academic practices, whereas later similarities point to the standardization and globalization of cartographic design in scholarly communication. In this sense, map design evolution appears less as a divergence between cultural traditions and more as a process of institutional convergence under shared publication systems.

## 6.2Implications for academic map design

Although statistically detectable cross-language differences are observed in several indicators, the overall design structures of Chinese- and English-language academic maps are highly similar. Across map elements, color usage, and layout structure, both groups exhibit restrained chromatic strategies, moderate alignment flexibility, strong centering tendencies, and increasing structural completeness over time. More importantly, the longitudinal trajectories are remarkably parallel: element richness, legend usage, and hue diversity all increase steadily in both contexts, while brightness contrast decreases and layout balance remains stable.

**(1) Color design**

Although English-language maps exhibit slightly higher saturation, hue richness ($N_{hue}$), and color entropy ($E_{hue}$), both language groups overwhelmingly rely on neutral dominant hues, low overall saturation, high brightness ($B_{ave}$), and restrained tonal contrast ($B_{con}$). For example, more than 85% of maps in both datasets use no more than four hues, and approximately 75% of maps remain concentrated in low-saturation and high-brightness ranges. More importantly, the temporal trajectories of all color indicators are highly synchronized. $N_{hue}$ and $E_{hue}$ increase steadily in both languages, $B_{con}$ declines, and overall brightness remains consistently high. These parallel trends suggest that academic color design is evolving along a shared path toward greater chromatic richness combined with moderated tonal differentiation. For contemporary map design practice, these findings imply that color strategies in academic cartography are largely standardized across linguistic contexts. Designers may confidently adopt restrained palette structures with moderate hue diversity, high luminance backgrounds, and controlled brightness contrast without needing language-specific adaptation.

**(2) Layout structure**

Layout indicators exhibit even stronger cross-language consistency. As shown in *Section* 5, hierarchy deviation ($D_{hier}$), compactness ($R_{map}$), alignment ($R_{horizontal}$ and $R_{vertical}$), and visual balance ($B_{horizontal}$ and $B_{vertical}$) display substantial distributional overlap between Chinese- and English-language maps. Although certain differences reach statistical significance (e.g., slightly higher $R_{map}$ in English maps), the magnitude of these gaps remains small relative to their full scales. Temporal analysis further reinforces this structural similarity. Across three decades, $R_{map}$ declines gradually in both groups—from approximately 0.82 to around 0.75—indicating a shared shift toward more information-dense layouts. Meanwhile, $D_{hier}$ stabilizes around 0.12, and visual balance indicators remain close to equilibrium in both contexts. Alignment measures fluctuate but exhibit no transformative divergence. These patterns suggest that academic layout design is governed by persistent compositional principles—centered main maps, high main-map occupation ratios, near-equilibrium balance, and moderate alignment flexibility—regardless of language. For practical map design, this indicates that academic cartography adheres to stable structural conventions rather than culturally divergent layout systems.

## 6.3 Limitations and future work

While this study provides a longitudinal and multilingual examination of thematic map design evolution, several limitations should be acknowledged.

First, the dataset is restricted to academic journal publications between 1990 and 2020, and the majority of analyzed cases follow a single–main-map layout structure. Although the selected journals represent authoritative sources in Chinese and English cartography, they do not encompass all forms of thematic maps. In particular, multi-main-map layouts, composite figures, and small-multiple designs have become increasingly common in recent years, especially in interdisciplinary and data-rich publications. The current study therefore does not explicitly model interactions among multiple coordinated map panels. As such, the findings mainly reflect conventions within single-map academic compositions rather than the full structural diversity of contemporary cartographic practice. Future research may extend the analytical framework to systematically examine multi-map configurations and explore whether similar convergence patterns persist in more complex layout systems.

Second, the quantitative indicators employed in this study focus on structural and chromatic dimensions, including element richness, hue diversity, hierarchical deviation, compactness, alignment, and balance. While these metrics capture important aspects of visual composition, other design dimensions (such as typography, symbol semantics, label density, annotation style, and thematic classification strategies) are not systematically measured. Future work could incorporate additional feature extraction methods to construct a more comprehensive representation of map design.

Third, although cross-language convergence is demonstrated across multiple indicators, the present analysis remains observational rather than causal. The study identifies synchronized trends but does not directly disentangle the relative influence of technological advances, editorial guidelines, software standardization, or broader globalization processes. Future research may combine policy analysis, editorial guideline comparison, or interviews with practitioners to further clarify the mechanisms underlying standardization.

## 7. Conclusion

This study investigates the longitudinal evolution of thematic map design in academic cartography and examines whether cross-language differences constitute enduring divergence or transitional variation. By constructing a multilingual dataset spanning three decades and operationalizing both color and layout characteristics into quantitative indicators, the study provides large-scale empirical evidence on how academic map design develops over time. The results indicate that cross-language differences are observable in structural completeness, chromatic richness, and certain layout metrics, particularly in the early stages of the study period. However, these differences do not mature into parallel design paradigms. Instead, most indicators exhibit synchronized temporal trajectories, and early gaps gradually diminish. Academic thematic maps in both Chinese- and English-language journals converge toward increasingly similar configurations characterized by moderate hue diversity, centered hierarchy, stable visual balance, and growing structural completeness. These findings suggest that contemporary academic cartography operates within an increasingly standardized visual system shaped by shared publishing infrastructures, digital production technologies, and institutional norms. Cultural context appears to influence design tendencies, yet long-term evolution is marked more by convergence than divergence.

## Acknowledgments

The authors would like to thank the editors and anonymous reviewers for their useful comments on the manuscript.

## Disclosure statement

No potential conflict of interest was reported by the author(s).

## Funding

This work was supported by grants from the National Natural Science Foundation of China (No. 42501551, 42371455), Tobii China Innovation Initiative Project (TPI250407CN).

## Data availability statement

The data that support the findings of this study are all openly available on Huggingface. The website: https://huggingface.co/datasets/song12321/papermap.

## Reference

Bertin, J. (1974). Graphische semiologie: diagramme, netze, karten. Walter de Gruyter.

Brewer, C. A. (1994). Color use guidelines for mapping. Visualization in modern cartography, 1994(123-148), 7.

Brewer, C. A., MacEachren, A. M., Pickle, L. W., & Herrmann, D. (1997). Mapping mortality: Evaluating color schemes for choropleth maps. Annals of the Association of American Geographers, 87(3), 411-438.

Brewer, C. A., Hatchard, G. W., & Harrower, M. A. (2003). ColorBrewer in print: a catalog of color schemes for maps. Cartography and geographic information science, 30(1), 5-32.

Brewer, C. A. (2016). Designing Better Maps: A Guide for GIS Users, Second Edition. Redlands, California: Esri Press.

Brychtova, A., & Coltekin, A. (2015). Discriminating classes of sequential and qualitative colour schemes. International Journal of Cartography, 1(1), 62-78.

Cao, Y., Yang, P., Xu, M., Li, M., Li, Y., & Guo, R. (2025). A novel method of urban landscape perception based on biological vision process. Landscape and Urban Planning, 254, 105246.

Christophe, S. (2011). Creative colours specification based on knowledge (ColorLegend System). The Cartographic Journal, 48(2), 138-145.

Coetzee, S., Carow, S., & Snyman, L. (2021). How are maps used in research? An exploratory review of PhD dissertations. Advances in Cartography and GIScience of the ICA, 3, 3.

Dayama, N. R., Todi, K., Saarelainen, T., & Oulasvirta, A. (2020, April). Grids: Interactive layout design with integer programming. In Proceedings of the 2020 CHI Conference on Human Factors in Computing Systems (pp. 1-13).

Dent, B. D., Torguson, J. S., & Hodler, T. W. (2008). Cartography: Thematic Map Design. 6th Edition. Boston: McGraw-Hill.

Harrower, M., & Brewer, C. A. (2003). ColorBrewer. org: an online tool for selecting colour schemes for maps. The Cartographic Journal, 40(1), 27-37.

He, B., Dong, W., Liao, H., Ying, Q., Shi, B., Liu, J., & Wang, Y. (2023). A geospatial image based eye movement dataset for cartography and GIS. Cartography and Geographic Information Science, 50(1), 96-111.

Huang, L., Chen, H., Wang, X., & Chen, G. (2000). A fast algorithm for mining association rules. Journal of Computer Science and Technology, 15(6), 619-624.

Isola, P., Zhu, J. Y., Zhou, T., & Efros, A. A. (2017). Image-to-image translation with conditional adversarial networks. In Proceedings of the IEEE conference on computer vision and pattern recognition (pp. 1125-1134).

Jia, F., Yang, J., Ding, L., Wang, G., & Song, G. (2024). An ontology-based semantic description model of ubiquitous map images. Transactions in GIS, 28(3), 457-485.

Kessler, F. C., & Slocum, T. A. (2011). Analysis of thematic maps published in two geographical journals in the twentieth century. Annals of the Association of American Geographers, 101(2), 292-317.

Kirillov, A., Mintun, E., Ravi, N., Mao, H., Rolland, C., Gustafson, L., ... & Girshick, R. (2023). Segment anything. In Proceedings of the IEEE/CVF international conference on computer vision (pp. 4015-4026).

Kramm, T., Nyamari, N., Moseti, V., Klee, A., Vehlken, L., Anderson, D. M., ... & Bareth, G. (2025). Deep learning-based extraction of Kenya's historical road network from topographic maps. Scientific Data, 12(1), 1149.

Li, W., & Hsu, C. Y. (2022). GeoAI for large-scale image analysis and machine vision: Recent progress of artificial intelligence in geography. ISPRS International Journal of Geo-Information, 11(7), 385.

Li, J., Yang, J., Hertzmann, A., Zhang, J., & Xu, T. (2019). Layoutgan: Generating graphic layouts with wireframe discriminators. arXiv preprint arXiv:1901.06767.

Jyothi, A. A., Durand, T., He, J., Sigal, L., & Mori, G. (2019). Layoutvae: Stochastic scene layout generation from a label set. In Proceedings of the IEEE/CVF International Conference on Computer Vision (pp. 9895-9904).

Ma, J., Wang, G., Cui, X., & Qi, X. (2013). Using the Principle of Moment Balance Establishing Map’ s Visual Balance Model. Geomatics and Information Science of Wuhan University, 38(1), 116-120.

MacEachren, A. M. (2004). How maps work: representation, visualization, and design. Guilford Press.

Mandal, S., Biswas, S., Das, A. K., & Chanda, B. (2014). Land Map Image Dataset: Ground-Truth And Classification Using Visual And Textural Features. Image Processing & Communications, 19(4), 37.

Mitrev, A. B., & Mirceva, G. (2025, April). YOLOv8 Oriented Bounding Box (OBB). In ICT Innovations 2024. TechConvergence: AI, Business, and Startup Synergy: 16th International Conference, ICT Innovations 2024, Ohrid, North Macedonia, September 28 – 30, 2024, Proceedings (Vol. 2436, p. 95). Springer Nature.

O'Donovan, P., Agarwala, A., & Hertzmann, A. (2014). Learning layouts for single-pagegraphic designs. IEEE transactions on visualization and computer graphics, 20(8), 1200-1213.

Petitpierre, R., Uhl, J. H., di Lenardo, I., & Kaplan, F. (2024). A fragment-based approach for computing the long-term visual evolution of historical maps. Humanities and Social Sciences Communications, 11(1), 1-18.

Petitpierre, R. (2025). Studying Maps at Scale: A Digital Investigation of Cartography and the Evolution of Figuration. arXiv preprint arXiv:2511.19538.

Qin, Z., & Li, Z. (2017). Grouping rules for Effective legend design. The Cartographic Journal, 54(1), 36-47.

Rosenfeld, A., & Pfaltz, J. L. (1966). Sequential operations in digital picture processing. Journal of the ACM (JACM), 13(4), 471-494.

Robinson, A. C., Demšar, U., Moore, A. B., Buckley, A., Jiang, B., Field, K., ... & Sluter, C. R. (2017). Geospatial big data and cartography: research challenges and opportunities for making maps that matter. International Journal of Cartography, 3(sup1), 32-60.

Saeedimoghaddam, M., & Stepinski, T. F. (2020). Automatic extraction of road intersection points from USGS historical map series using deep convolutional neural networks. International Journal of Geographical Information Science, 34(5), 947-968.

Silva, S., Santos, B. S., & Madeira, J. (2011). Using color in visualization: A survey. Computers & Graphics, 35(2), 320-333.

Slocum T. A., McMaster, R. B., Kessler, F. C., & Howard, H. H. (2009). Thematic Cartography and Geographic Visualization (3rd edition). Upper Saddle River, NJ: Pearson/Prentice Hall.

Tait, A. (2018). Visual Hierarchy and Layout. The Geographic Information Science & Technology Body of Knowledge (2nd Quarter 2018 Edition), John P. Wilson (ed.). Website: https://gistbok-topics.ucgis.org/CV-03-007

Ware, C. (2019). Information visualization: perception for design. Morgan Kaufmann.

Wang, L., Zhang, J., Weng, M., Kang, M., & Su, S. (2024). Unlocking semantic information representation in bar graph design. IEEE Transactions on Visualization and Computer Graphics.

Wang, B., Xu, C., Zhao, X., Ouyang, L., Wu, F., Zhao, Z., ... & He, C. (2024). Mineru: An open-source solution for precise document content extraction. arXiv preprint arXiv:2409.18839.

Wei, J., Guo, Q., Wei, Z., Liu, Q., & Liu, Y. (2017). On the method of automatic thematic map layout. Engineering of Surveying and Mapping, 26(10), 12-17.

White, T. M., Slocum, T. A., & McDermott, D. (2017). Trends and issues in the use of quantitative color schemes in refereed journals. Annals of the American Association of Geographers, 107(4), 829-848.

Wu, M., Lv, G., Qiao, L., Roth, R. E., & Zhu, A. X. (2024). Green Cartography: A research agenda towards sustainable development. Annals of GIS, 30(1), 15-34.

Wu, C., Li, J., Zhou, J., Lin, J., Gao, K., Yan, K., ... & Liu, Z. (2025). Qwen-image technical report. arXiv preprint arXiv:2508.02324.

Yang, N., Wang, Y., Wei, Z., & Wu, F. (2025). MapColorAI: designing contextually relevant choropleth map color schemes using a large language model. Cartography and Geographic Information Science, 1-19.

Yang, Z., Zhang, H., Wang, J., & Li, M. (2025). MapLayNet: Map layout representation learning using weakly supervised structure-aware graph neural networks. International Journal of Geographical Information Science, 39(2), 201-225.

Zhang, Y., Chen, X., Zheng, W., Guo, Y., Li, G., Chen, S., & Yuan, X. (2025). VisTaxa: Developing a Taxonomy of Historical Visualizations. IEEE Transactions on Visualization and Computer Graphics.

Zhou, X., Li, W., Arundel, S. T., & Liu, J. (2018). Deep convolutional neural networks for map-type classification. arXiv preprint arXiv:1805.10402.

Zhou, X., Wen, Y., Shao, Z., Li, W., Li, K., Li, H., ... & Yan, Z. (2024). CartoMark: a benchmark dataset for map pattern recognition and map content retrieval with machine intelligence. Scientific Data, 11(1), 1205.